\documentclass[runningheads]{llncs}
\usepackage[T1]{fontenc}
\usepackage{graphicx}
\usepackage{booktabs}
\usepackage[misc]{ifsym}
\usepackage{amsmath}
\newcommand{\corr}{(\Letter)}
\usepackage{xcolor}
\usepackage[normalem]{ulem} 
\graphicspath{ {figures/} }
\begin{document}
\title{Stop Guessing: Optimizing Goalkeeper Policies for Soccer Penalty Kicks}

\titlerunning{Stop Guessing: Optimizing Goalkeeper Policies for Soccer Penalty Kicks}

\author{Lotte Bransen\inst{1} \corr \orcidID{0000-0002-0612-7999} \and
Tim Janssen\inst{2}  \and
Jesse Davis\inst{1}}

\authorrunning{L. Bransen et al.}

\institute{KU Leuven \email{\{lotte.bransen, jesse.davis\}@kuleuven.be}
\and
KNVB \email{tim.janssen@knvb.nl}}

\maketitle

\begin{abstract}
Penalties are fraught and game-changing moments in soccer games that teams explicitly prepare for. Consequently, there has been substantial interest in analyzing them in order to provide advice to practitioners. From a data science perspective, such analyses suffer from a significant limitation: they make the unrealistic simplifying assumption that goalkeepers and takers select their action -- where to dive and where to the place the kick -- independently of each other. In reality, the choices that some goalkeepers make depend on the taker's movements and vice-versa. This adds substantial complexity to the problem because not all players have the same action capacities, that is, only some players are capable of basing their decisions on their opponent's movements. However, the small sample sizes on the player level mean that one may have limited insights into a specific opponent’s capacities. We address these challenges by developing a player-agnostic simulation framework that can evaluate the efficacy of different goalkeeper strategies. It considers a rich set of choices and incorporates information about a goalkeeper's skills. Our work is grounded in a large dataset of penalties that were annotated by penalty experts and include aspects of both kicker and goalkeeper strategies. We show how our framework can be used to optimize goalkeeper policies in real-world situations.


\end{abstract}

\section{Introduction}

Penalty kicks in soccer often are associated with high-leverage moments.  On the one hand, 65\% of the penalties are awarded in tied games or when a team is trailing by a goal. Because penalties are converted at substantially higher rates than open-play shots (78\% vs. 12\%), these represent a significant chance to alter a game’s outcome. On the other hand, even though penalty shootouts are only used to decide the winner of games that are tied after extra time in the knockout stages of tournaments such as the FIFA World Cup and the UEFA Champions League, the low-scoring nature of soccer means that penalty shootouts are surprisingly common. Since 1990, 23\% of the men's World Cup knockout games, 27\% of the men's Euros knockout games, and 23\% of the men’s Champions League’s finals have ended in a shootout. 

Consequently, teams explicitly prepare for penalties. For example, Louis van Gaal famously substituted in goalkeeper Tim Krul prior
to the Netherlands’ shootout against Costa Rica in the quarterfinals of the 2014 World Cup. Krul saved two kicks and the Netherlands advanced. More recently, at the 2024 European Championship, English goalkeeper Jordan Pickford had a list taped to his waterbottle with instructions for each potential Swiss taker.\footnote{It is unclear how this list was created and what role data played. For an image, see: \url{https://news.sky.com/story/england-keeper-jordan-pickfords-not-so-secret-weapon-during-euro-2024-penalty-shootout-victory-over-switzerland-13174555}} The instructions for the first taker, Manuel Akanji, said to dive left. He did and saved the kick.

Research has investigated penalties from a variety of different perspectives. 
Psychological studies seek to understand the mental aspects of penalties, particularly related to stress. They aim to develop strategies to prepare players to cope with pressure~\cite{palacioshuerta2010,jordet2012b,Wood:lottery}. 
Performance analysis studies have looked at how to support the kicker by investigating where a kick should be placed to optimize the chance of converting~\cite{horn2020,almeida2016,bareliazar2009}. These consider aspects such as the kicking technique and speed~\cite{hunter2022}.  
Finally, researchers have studied the choices that players make from a game-theoretic perspective~\cite{palacioshuerta2003,tuyls2021}. Interestingly, in aggregate, players closely follow a mixed-Nash
equilibrium strategy. However, partitioning the takers based on playing styles leads to different recommendations~\cite{tuyls2021}. 

Unfortunately, the existing work has several important limitations. First, the game-theoretic approaches have greatly simplified the problem by assuming that both players \emph{independently pre-select}  their decision~\cite{palacioshuerta2003,tuyls2021}. This induces a major misalignment with observed strategies where some takers and goalkeepers \emph{base their decision on each other's movements}. That is, some takers employ a \emph{keeper-dependent} strategy in which they select where to place their kick by reacting to the goalkeeper's behavior \cite{noel2014} whereas some goalkeepers delay their dive in order to base its direction on the kicker's movements \cite{noel2021}. However, we will show that simply incorporating these aspects into a game-theoretic approach does not work because not all players have the same action capacities (i.e., what physical actions a player is able to perform)~\cite{fajen2007}. That is, only some players are capable of basing their decisions on an opponent's movements and the small sample sizes means that one may have limited insights into an opponent’s capacities. Second, little work analyzes the tactical choices available to the goalkeeper such as providing advice about their initial positioning (e.g., should they stand in the center of the goal or shade themselves slightly to one side). 

We overcome these weaknesses by proposing a simulation-based approach for analyzing the efficacy of different goalkeeper strategies. Our approach confers several important advantages. First, it considers a rich set of tactical choices that others do not, including whether or not the goalkeeper bases their dive on the taker's movements, and the goalkeeper's initial position. Second, the approach can be configured based on the specific action capacities of the goalkeeper, which can be measured in a testing or training environment.  Third, it addresses sample-size issues about takers by learning player-agnostic models. Finally, our study is grounded in a large dataset of 7,872 penalties where penalty experts annotated, e.g., the strategies employed by both takers and goalkeepers. Consequently, the framework allows for a coaching staff to provide tailored advice based on a specific goalkeeper's capabilities. Moreover, it can support making informed decisions (e.g., whether to substitute in a goalkeeper for a penalty shootout). Our experiments show that the best policy for a specific goalkeeper depends on their action capacities and aligning slightly off center increases the chance of saving a kick.

\section{Preliminaries}

Penalty kicks in soccer are taken 11 meters (12 yards) from the center of the goal and only involve the kicker and the goalkeeper. The rules stipulate that (1) {\bf the goalkeeper} must stand with one foot on or behind the goal line until the ball has been kicked; and (2) {\bf the kicker} must not feint or hesitate while running up to kick the ball~\cite{fifalawsofthegame}.

When certain games in knockout competitions end in a tie, a penalty shootout determines the winner. Penalty shootouts start with a coin flip to decide which team kicks first and on which goal the kicks will be taken. The teams alternate taking penalties. Initially, a shootout involves 10 penalties and the teams the converts the most penalties wins. If the score is tied after 10 penalties, the shootout enters sudden death: in each round a player from both teams takes a kick. This continues until there is a round in which one team scores and the other does not. A player can only take a second penalty kick after all their teammates have taken a kick. Teams typically determine the order of takers prior to the shootout. Often, experienced penalty takers attempt the first kicks and less-experienced players take later kicks (see Appendix \ref{app:order-experience}).

\subsection{Penalties from a Takers' Perspective}

Since players tend to kick the ball with the inside of their foot, the kicking motion makes it more \emph{natural} and thus more common for right-footed (left-footed) players to kick toward the left (right) \cite{palacioshuerta2003}.
Thus, instead of referring to the left and right corners of the goal, the convention is to refer to the natural and non-natural corners, based on the kicker's dominant foot. 

When taking a penalty, kickers tend to employ one of two strategies:
\begin{description}
\item[Keeper-independent] kickers predetermine the location they will aim the kick toward. Thus, the direction is selected independent of the goalkeeper's movements. 
\item[Keeper-dependent] kickers watch the goalkeeper’s movements and select the kick direction based on those movements \cite{kuhn1988penalty}.
\end{description}
Most players follow the independent strategy. The dependent  approach is harder and can degrade performance because the kicker has a limited amount of time to modify their kicking motion~\cite{vanderkamp2006}. Moreover, its success is also tied to the goalkeepers’ temporal decision-making \cite{noel2021}.

\subsection{Penalties from a Goalkeepers' Perspective}
The goalkeeper's primary decisions revolve around where and when to dive. The short distance between the ball and the goal makes it challenging for a goalkeeper to react to the kick \cite{dicks2010b,savelbergh2005} as ball flight times vary between 500ms and 600ms \cite{dicks2010a,zheng2021}. Beyond this, a goalkeeper's dive range (i.e., how far they can dive in a set amount of time) depends on their physical characteristics. This range varies among goalkeepers. 

We distinguish between two different strategies a goalkeeper can employ:
\begin{description}
\item[Early dive] goalkeepers guess a direction and start moving toward one side of the goal shortly before the shooter kicks the ball. 
\item[Late dive] goalkeepers  base their dive direction on the kickers movement by delaying their movement to collect more and more accurate information from the kicker to increase their chances of diving to the correct location~\cite{hunter2018b}. 
\end{description}
There is a trade-off between these two strategies. The early dive strategy affords the goalkeeper more time to move, meaning that they have a larger dive range (i.e., more likely to reach the corners) but have less information about the kicker's intentions. A late dive allows the goalkeeper to glean information from the kicker's run up that might help them dive toward the correct corner, but they will have a shorter dive range due to having less time. The footnote links to illustrative videos showing this trade-off.\footnote{An early dive \url{https://youtu.be/OjAUTduO\_Fk?t=70}  and a late dive \url{https://youtu.be/OjAUTduO\_Fk?t=142} vs. a keeper-dependent kick. When the goalkeeper dives early, the kicker chooses the other corner. On the late dive, the goalkeeper picks the correct corner, but fails to reach the well-placed kick.}

\subsection{Data}

We had access to two data sources that provided disjoint information. First, a penalty expert annotated kicks of players likely take part in the 2022 FIFA Men's World Cup. Most importantly, this data contains information about the kicker's strategy (keeper-independent or keeper-dependent). It also recorded the goalkeeper's dive strategy for about half of the kicks. Second, we had event data as collected by commercial data providers about most matches.  This does not record the kicker's or goalkeeper's strategy. However, it does record three important things  not present in first dataset: (1) the end coordinates (x,y,z) of the penalty kick, either where it was stopped by the goalkeeper or where it passes the goal line, (2) information about what every player did during the game (e.g., position played, actions executed), and (3) the order of penalty takers in a shootout, which cannot be derived from first dataset since only includes kicks involving a player likely to participate in the 2022 World Cup. 

Because they contain unique information needed for analyzing penalties, we merge the datasets.  However, this requires matching penalties between the two datasets. This is not straightforward as they were collected in different ways by different organizations.  Hence, relevant information such as team names (e.g., Athletic Bilbao versus Athletic Club), player names (e.g. Sami Al Naji versus Halil Al Naji Sami), and times (e.g., timezone used, representation of the date) are recorded differently. Moreover, the datasets do not contain the same set of matches. Appendix \ref{app:merging_data} describes our procedure to merge the data. 

After merging, we had 7,872 kicks, including 1,201 taken in a shootout. These penalty kicks involve 722 unique penalty takers and 161 distinct goalkeepers. To highlight the sample size issues, only 59 goalkeepers and 28 takers were involved in at least 25 kicks.  The average conversion rate of penalty kicks in our dataset is 77.9\%. Players place the kick on target 93.5\% of the time (i.e., 6.5\% of the kicks are wide or high). In terms of taker's strategy, 20.6\% kicks were taken keeper-dependently and 79.4\% were taken keeper-independently. We define a goalkeeper's save percentage as the number of kicks saved over the number of kicks on target. On average goalkeepers save 16.8\% of the kicks on target. 
The goalkeeper's dive timing was initially not recorded and is only available for 3,997 kicks. In these kicks, goalkeepers use a late dive on 38.5\% of penalties and an early dive 61.5\% of the time.

\section{Penalty Game Theory with Player Tactics}
To motivate our approach, we will start with the typical game-theoretic approach that views a penalty kick as a zero-sum game between the kicker and the goalkeeper \cite{palacioshuerta2003,tuyls2021}. Whereas prior work simplifies the setting by assuming that all takers use the independent strategy and all goalkeepers dive early, we will consider a richer setting where each player has a novel choice: the goalkeeper can decide to dive late and the kicker can take the kick keeper-dependently. We will see that the game-theoretic approach is less well suited to modeling these strategies due to differences in players' action capacities. 

In our augmented setting, the kicker has four choices: kick to natural corner (N), non-natural corner (NN), center (C) and kick dependently (Dep). The goalkeeper has three choices: diving early toward the kicker's natural corner, diving toward the non-natural corner or diving late.\footnote{An early dive to the center is not considered a choice because this is indistinguishable from staying in the center on a late dive when the kicker does not aim for a corner.} The players' choices should be made simultaneously.

\begin{table}
\caption{The empirical payoff matrix for the Penalty Kick on our dataset.}
\begin{center}
\begin{tabular}{l|ccc}
                  & \textbf{GK N} & \textbf{GK Late}     & \textbf{GK NN}    \\ \hline
\textbf{Kick N}   & 0.615         & 0.785            & 0.939          \\
\textbf{Kick C}   & 0.846         & 0.273            & 0.865          \\
\textbf{Kick NN}  & 0.947         & 0.785            & 0.556          \\
\textbf{Kick Dep} & 0.846         & 0.773            & 0.846         
\end{tabular}
    
\end{center}
\label{tbl:payoff}
\end{table}

Table \ref{tbl:payoff} shows the payoff matrix for the Penalty Kick game as estimated empirically from our dataset. We used the Minimax theorem \cite{vonneumann1947} to find the optimal mixed strategies for both players: 

\begin{itemize}
    \item \textbf{Goalkeepers: (N, Late, NN)} = (6.9\%, 87.1\%, 6.0\%)
    \item \textbf{Kickers: (N, C, NN, Dep)} = (43.1\%, 0.0\%, 35.7\%, 21.1\%)
\end{itemize}

Goalkeepers should dive late 87.1\% of the time and dive early 12.9\% of the time (6.9\% to the natural corner and 6.0\% to the non-natural corner). Kickers should employ the keeper-dependent strategy 21.1\% of the time and the independent strategy 78.9\% of the time (natural corner 43.1\%, center 0\%, and non-natural corner 35.7\%). Note that one should not perform a keeper-independent kick to the center: if a taker wants to aim in this zone, then they should use the dependent strategy to ensure that the goalkeeper dives towards a corner (i.e., vacates the central zone) prior to aiming there. 

In contrast to what has been observed in the simplified setting~\cite{palacioshuerta2003,tuyls2021}, we see a stronger divergence from the optimal mixed strategy with players empirically following:
\begin{itemize}
    \item \textbf{Goalkeepers: (N, Late, NN)} = (35.9\%, 38.5\%, 25.6\%)
    \item \textbf{Kickers: (N, C, NN, Dep)} = (39.5\%, 11.5\%, 28.4\%, 20.6\%)
\end{itemize}
\noindent In particular, we see two substantial deviations as (1) goalkeepers dive late much less often in reality than the optimal strategy suggests they should (38.5\% vs. 87.1\%), and (2) takers aim a keeper-independent kick to the center 11.5\% of the time, whereas the optimal strategy is never to do this. 

The mismatch between observed behavior and optimal behavior likely arises due to the differing action capacities of the players. Not all goalkeepers have the action capacity to execute the late dive strategy~\cite{hunter2018b} and even if they can, their range on a late dive will vary~\cite{zheng2021}. Similarly,  not all takers have the action capacity to take goalkeeper-dependent kicks ~\cite{vanderkamp2006}. Extending the analysis is challenging, because there is substantial uncertainty about players' action capacities due to the small sample sizes and players' evolving skill sets. Moreover, it is possible that the empirical conversion rates are biased because, e.g., we have more observations about players with these capacities. Therefore, these values may not generalize, which impacts situations such as shootouts that are likely to involve less-experienced takers. Consequently, a more nuanced approach is needed that can consider both  action capacities and contextual factors in a more player-agnostic way. 

\section{Modeling Goalkeeper Policies for Penalty Kicks}

We propose a simulation framework that makes use of learned models and allows us to better account for the variability of goalkeepers' action capacities.
A goalkeeper's strategy involves three major components. First, the goalkeeper can decide where to lineup with respect to the goal line. For example, they may lineup in the center or may want to shade themselves slightly toward one corner. Second, they can decide to dive early or late. Third, they need to decide which direction to dive, which depends on the previous point. We refer to a setting for each of these choices as a policy $\pi$. Formally, we address the following task:
\begin{description}
\item[Given:] A goalkeeper's capacities, a policy $\pi$, and historical penalties
\item[Do:] Compute what percentage of on-target penalties the goalkeeper would stop when following $\pi$.
\end{description}
\noindent We address this task by estimating the probability of saving penalty kick $k$ as: 
\begin{equation}
    p(s|k,\pi,{gk}) = p(s|k,c,\pi,{gk}) \times p(c|k,\pi,{gk})  
\end{equation}
where  $p(c|k,\pi,{gk})$ is the probability to choose the correct corner for penalty $k$ when goalkeeper ${gk}$ uses policy $\pi$,  and $p(s|k, c,\pi,{gk})$  is the probability to stop penalty $k$ given that you dive to the correct corner $c$ using policy $\pi$ for goalkeeper ${gk}$.  Instead of trying to simulate an end location of a penalty, we will evaluate policies on the actual end location of historical penalties (i.e., the x,y,z coordinate), though care will be taken when a kicker employs the keeper-dependent strategy.\footnote{Given that keeper-dependent kickers choose their direction based on the goalkeeper's movements, we cannot use the actual x,y,z coordinate to evaluate goalkeeper policies.} Next, we describe how to design informed policies and compute these two probabilities. 

\subsection{Specifying Educated Policies with Learned Models}
When specifying dive timing and dive direction policies, a natural question is whether it is possible to use learned models to make a more informed decision. Such models could be used in two important ways within a policy. First, having a better indication of what zone the ball will be directed toward can allow a goalkeeper to make an educated guess when executing an early dive. Second, knowing how far away from the goalkeeper's position the ball will be placed can help a goalkeeper decide whether to dive early or late: if a kick is likely to be out of a late goalkeeper's reach, then they should dive early.

These are challenging problems to solve due to small sample sizes of historical kicks for a given taker and the fact that game context features can affect these choices. Therefore, we use generic information about the takers and the game context. Moreover, to remain flexible, these models do not depend on knowing the goalkeepers diving range, which is typically only known to their team. Specifically, we describe each penalty using 47 features that can be broken down into five categories: contextual features (e.g., time in the game, goal difference), player characteristics (e.g., preferred foot, position), penalty taking experience (e.g., number of taken penalties), penalty taker's history (e.g., percentage of previous kicks to natural and non-natural corners, average distance that ball is placed from the center of the goal), and
shootout features (e.g., result and direction of the last penalty kick taken by the opponent and by the kicker's own team). These are described in more detail in Appendix~\ref{sec:app-features}. 

We train two models:
\begin{description}
    \item[Penalty Direction Model] predicts whether the kick will be aimed toward the (1) center, (2) the kicker's natural corner or (3) the kicker's non-natural corner. Specifically, we train an XGBoost \cite{chen2016} multi-class probablistic classifier on the keeper-independent kicks in our dataset.
    \item[Penalty Distance Model] predicts the distance of the end location of the kick to the center of the goal. Specifically, we train an XGBoost regressor.
\end{description}

We consider four different policies for dive timing and direction. In the \textbf{late} policy, the goalkeeper always dives late. In the \textbf{early} policy, the goalkeeper always dives early. The direction is selected according to the observed probability of diving to each zone (i.e., this policy does not use a learned model). In the \textbf{early educated} policy, the goalkeeper always dives early but dives toward each zone with a weighted probability as predicted by the learned Penalty Direction Model. Finally, in the \textbf{mixed educated} policy the goalkeeper bases the decision about whether to dive early or late on the learned Penalty Distance Model. If the predicted distance is within the goalkeeper's late dive range, then they will dive late. Otherwise, they use the early educated policy. 

\subsection{Computing $p(c|k,\pi,{gk})$} 
The probability to choose the correct corner depends on both the goalkeeper's dive type and the taker's strategy. When a goalkeeper's policy involves an early dive against a keeper-dependent taker, the kicker will choose an opposing corner to aim for. Hence, the goalkeeper is unlikely to guess the correct corner.\footnote{The taker could mishit the ball, leading to a correct guess.} Therefore, we need to learn a separate probability for a combination of the four considered goalkeeper policies and the two taker strategies.  

For the \textbf{late} and \textbf{early} policies, we estimate the probabilities of guessing correctly empirically (i.e., by counting) from the merged dataset. For the \textbf{educated} policies, the direction of the early dive is determined by the Penalty Direction Model. Specifically, we sample a direction according to the distribution of the zones that are predicted by the model. Therefore, when facing an independent taker, the probability of guessing the correct corner when using an educated early dive is the probability predicted by the Penalty Direction Model for the kick's true location (i.e., observed end coordinates in the data).
Dependent kickers base their penalty direction on the goalkeeper's movement. Therefore, for dependent kicks, we cannot use the actual direction of the kick. Hence, we use the emperical value of choosing the correct corner when facing a dependent kick. This is computed over all dependent kicks in the dataset.

\subsection{Computing $p(s|k,c,\pi,{gk})$} 
To stop a shot, a goalkeeper must be able to \emph{reach} the point where the ball crosses the goal line \emph{prior} to it fully crossing the line. What locations a goalkeeper can reach depend on a number of factors such as when they start moving, how fast they move, and the speed of the ball being kicked. We will focus on the first factor, as when they start moving depends on their timing strategy. We use $r(\pi,{gk})$ to denote the goalkeepers dive range for the dive type (early or late) selected by policy $\pi$ for a specific kick.

Because we use historical kicks with known end coordinates, we could naively assume that $p(s|k,c,\pi,{gk})=1$ if the goalkeeper guesses the correct corner and $d(loc_s,k_e) < r(\pi,{gk})$ and 0 otherwise. Here, $d(loc_s,k_e)$ is the Euclidean distance between the goalkeepers starting location $loc_s$\footnote{Taken as the point on the goalline that is equidistant between their two feet} and the location $k_e$ where the ball crosses the goal plane.\footnote{Both locations are three dimensional}   
However, two issues arise: a) a goalkeeper will not stop all penalty kicks within reach because they may make an error in their technique or not be able to deflect the ball out of the goal, and b) the end coordinates are annotated manually which will make them less accurate. Therefore, we introduce some uncertainty and compute the probability that goalkeeper ${gk}$ saves penalty kick $k$ when they toward the correct corner as

\small
\begin{equation}
  p(s|k, c,\pi,{gk}) = \left\{
  \begin{array}{@{}ll@{}}
   0, & d(loc_s, k_e)(\pi) >  r(\pi,{gk}) + \mu\\
    \rho, &  d(loc_s, k_e)(\pi) <  r(\pi,{gk}) -\mu\\
    \rho \dot (\frac{1}{2} - \frac{d(loc_s, k_e)(\pi)-r(\pi,{gk})}{2\mu}), & \text{otherwise}
  \end{array}\right.
\end{equation}
\normalsize

\noindent If the kick ends within $\mu$ meters of the keeper's reach, the probability of stopping the ball linearly increases as ball's end location nears the goalkeeper initial position. $\rho$ represents the probability that a goalkeeper saves a penalty that they reach (e.g., they may fail due to a technique error). Figure \ref{fig:influence-rho-mu} in Appendix \ref{app:uncertainty} shows the influence of $\rho$ and $\mu$ on the save probability. 

\section{Evaluation}

The evaluation's goal is to (1) explore the efficacy of each policy for different diving ranges given the standard center of the goal initial alignment, (2) investigate the effect of the goalkeeper's initial position, (3) evaluate the quality of the individual learned models,  and (4) show illustrative practical use cases.

For the first two questions, we simulate each penalty in our dataset and compute the expected save percentage using Equation~1. We set $\mu=0.7m$ and $\rho=0.7$; these values were derived empirically as described in Appendix~\ref{app:uncertainty}.  We compare the four policies defined in Section 4.1: the \textbf{late}, \textbf{early}, \textbf{early educated} and \textbf{mixed educated} policy. Because we are unaware of other approaches that evaluate goalkeeper policies for soccer penalties, we compare our approach to the policy of following the game-theoretic optimum as derived in Section 3. We also report the empirical save percentage.

\subsection{Evaluating Goalkeeper Policies}

\begin{figure}[t]
\includegraphics[width=10cm]{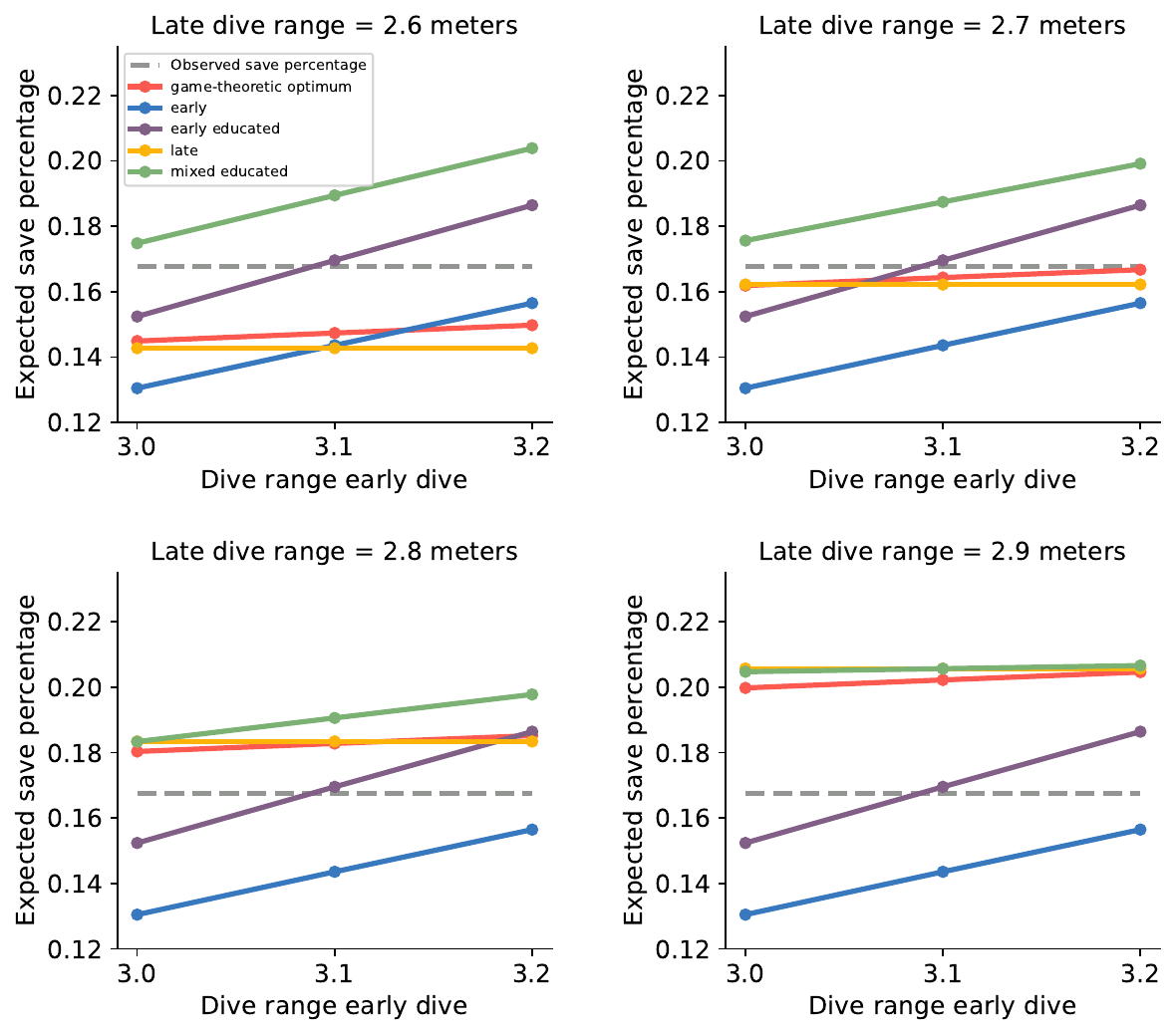}
\centering
\caption{Each plot shows how the expected save percentage varies as function of the early dive range for a fixed late dive range.}
\label{fig:experiments}
\end{figure}

We evaluate the expected save percentage for each policy as a function of a goalkeeper's dive ranges. Based on estimates from experts, we vary the late range from $2.6m$ to $2.9m$ and the early range from $3.0m$ to $3.2m$ in increments of $0.1m$. Figure \ref{fig:experiments} shows four plots, each showing how the expected save percentage varies as function of the early dive range for a fixed late dive range. 

The best policy depends on a goalkeeper's capacities. If the goalkeeper is capable of diving late, then the \textbf{mixed educated} policy is always the optimal policy. However, if the goalkeeper does not have the capacity to dive late, the \textbf{early educated} policy outperforms the \textbf{early} policy. When a goalkeeper has a lower late dive range, the game-theoretic optimum is outperformed by the \textbf{early educated} policy. This again illustrates the importance of incorporating action capacities into a framework in order to provide meaningful advice. Finally, the game-theoretic optimum closely tracks the late dive policy, as it states that the goalkeeper should dive late 87.1\% of the time. 

\subsection{Initial Goalkeeper Position}
Over time, the number of keeper-independent kicks placed close to the corners of the goal has increased: In 2021, 10.8\% of 1,140 such kicks were placed within 25cm of a post compared to only 5.0\% of such 524 kicks in 2016.  Such kicks are simply out of range (i.e., impossible to save) for any realistic early dive range \emph{if} the goalkeeper's starting position is in the  middle of the goal.  Hence, we investigate a policy where the goalkeeper's initial alignment is slightly closer to one corner. 
Research has indicated that player's are able to shoot more forcefully (i.e., higher ball velocity) when aiming toward their natural corner \cite{Lees1998TheBO}. Moreover, the natural corner is the most common placement location \cite{noel2014}. Therefore, we evaluate the efficacy of the goalkeeper lining up $0.1m$, $0.2m$, and $0.3m$ closer to the natural corner than the non-natural one.

\begin{figure}[t]
\includegraphics[width=6cm]{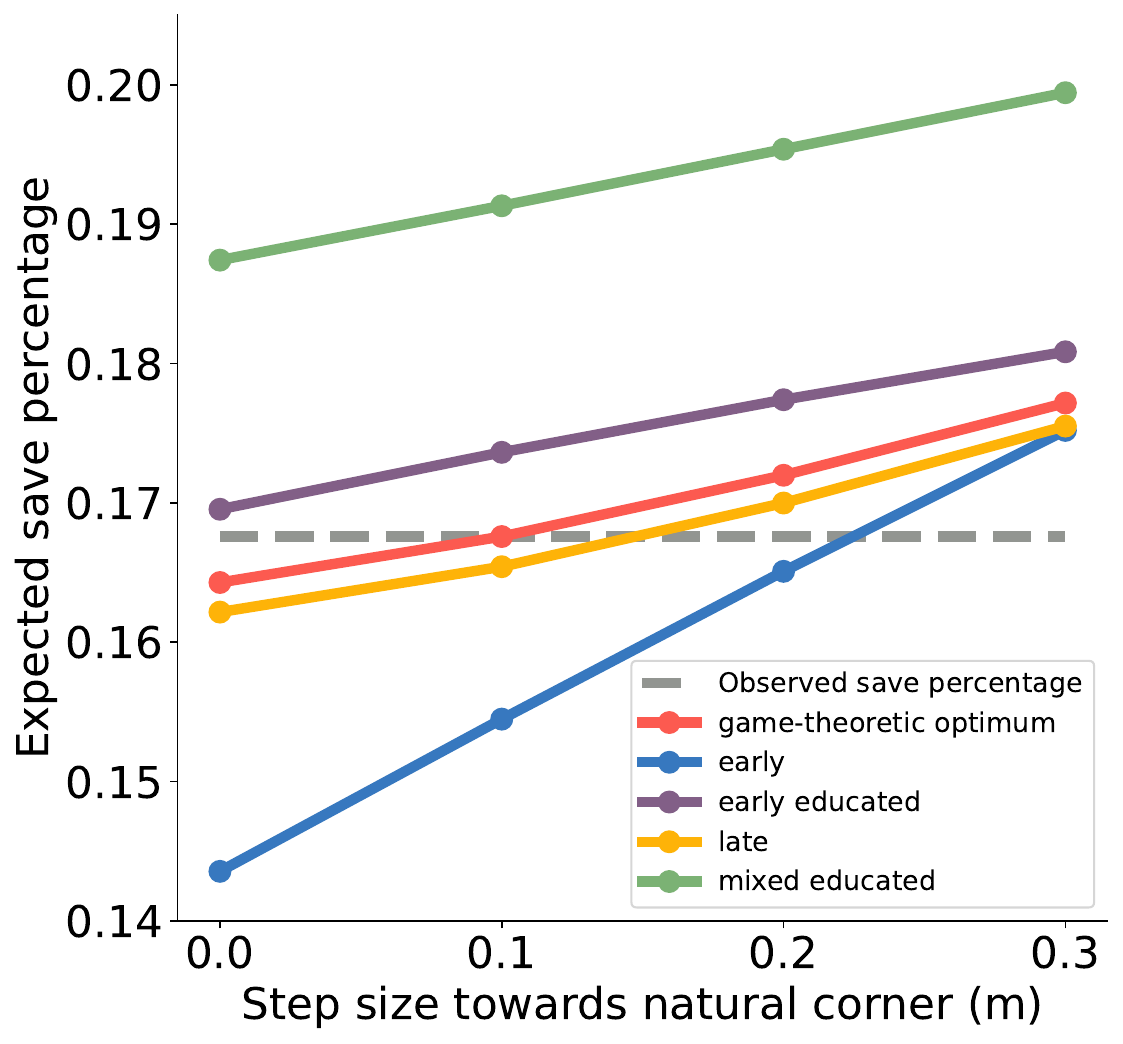}
\centering
\caption{Expected save percentage for on-target penalties as function of moving x meters toward the kicker's natural corner. 
}
\label{fig:step-to-the-side}
\end{figure}

Figure~\ref{fig:step-to-the-side} shows the estimated save percentage for a goalkeeper with an early dive range of $3.1m$ and a late dive range of $2.7m$ for each of our four policies as a function of their initial position. The results indicate that shading toward the natural corner can potentially increase the expected save percentage. However, players may change their penalty direction based on the positioning of the goalkeeper. Research shows that this might especially be the case when the goalkeeper is (clearly) not positioned in the center of the goal before the player’s run up to the ball \cite{masters2007,weigelt2012,noel2015}. Therefore, it may be advisable to alter one's alignment during the kicker's run toward the ball. As Figure \ref{fig:step-to-the-side} shows, this policy could increase the chance of saving a penalty by between 0.5 and 3 percentage points depending on which policy is used and how much one moves towards the kicker's natural corner. 

\subsection{Evaluating the Learned Models}
We use nested 5-fold cross-validation to train and evaluate our direction predictor and distance estimator models. We ensure that all penalties taken by the same player are in the same fold. We tune the max tree depth, learning rate, number of estimators and the minimum child weight (see Appendix~\ref{app:hp-set}).  

{\bf Predicting the Penalty direction.} We compare our learned classifier to the base model of just predicting the empirical distribution of aiming to each of the three considered zones. As we are interested in how well calibrated the probability estimates are, we report the logloss. On all penalties, our model obtains a logloss of $0.954 \pm 0.011 $ vs. $0.969 \pm 0.014$ for the base model. When only considering shootout penalties, our model obtains a logloss of $0.941 \pm 0.032$ whereas the base model's is $0.988 \pm 0.034$. The model is well calibrated; see Figure~\ref{fig:direction-calibration} in Appendix \ref{app:results}. 

\begin{figure}[t]
\includegraphics[width=9cm]{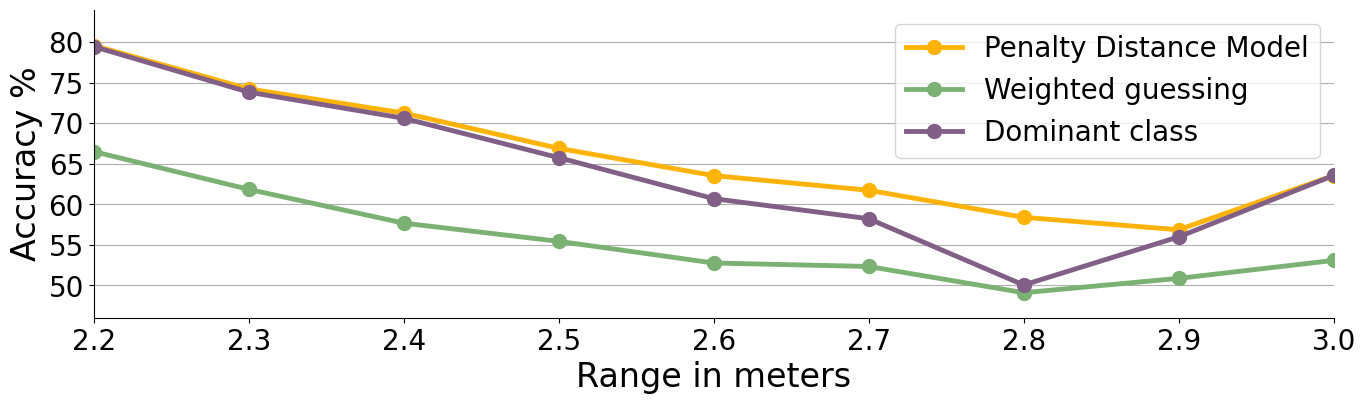}
\centering
\caption{Accuracy per range threshold for our Penalty Distance Model and baselines. Most shots are placed within 2.5 and 3 meters from the center, and this is also the area in which our model slightly outperforms the baselines.}
\label{fig:evaluation-distance}
\end{figure}

{\bf Predicting penalty distance.} We will use this model by thresholding its predicted value based on a goalkeeper's maximum range such that they can decide whether to dive early or late. Therefore, we vary this range and compute each model's accuracy. A prediction is correct if both the predicted and actual distance fall on the same side of the threshold. For each threshold, we compare our model's performance to two baselines: 1) the training set's average distance to the goalkeeper, and 2) randomly guessing whether the shot is within reach based on the training set's base rate for the specific threshold.
Figure \ref{fig:evaluation-distance} shows how the accuracy varies as a function of ranges. Our model slightly outperforms the baselines in the ranges between 2.5 to 2.9 meters from the center of the goal.

Appendix \ref{app:feature-insights} provides insights in what our models learned from the data and which feature combinations impact the predictions.

\subsection{Illustrative Use Cases}
We show three ways in which our proposed methodology could aid practitioners. In the first two use cases, we compare three goalkeepers with different action capacities. \textbf{GK1} has an average late diving reach ($2.8m$), average early diving reach ($3.1m$), and an average correct corner guess percentage when diving late ($p=0.59$). \textbf{GK2} has an above average early dive ($3.1m$) range, but is not able to dive late. Finally, \textbf{GK3} has lower diving ranges ($2.7m$ late range and $3.0m$ early range), but has a strong skill in choosing the correct corner based on the kicker's movement when diving late ($p=0.7$). The third use case discusses how our model could be used to provide player-specific advice to a goalkeeper.

\paragraph{UC1: How do action capacities affect a goalkeeper's optimal policy?}
For each of these goalkeepers we want to find their optimal policy for in-game penalty kicks (so excluding shootout kicks) given their capacities. Table \ref{tbl:use-cases} shows the expected save percentage for on target in-game penalty kicks for each goalkeeper for the three policies: \textbf{early educated}, \textbf{late} and \textbf{mixed}. In this use case we do not employ the tactic of taking a step to the side. We observe that each goalkeeper has a different optimal policy which is due to the differences in their action capacities. GK2 should employ the \textbf{early educated} policy, GK3 should employ the \textbf{late} policy and GK1 should employ the \textbf{mixed} strategy. When these goalkeepers employ their best strategy, GK3 (19.4\%) and GK1 (19.3\%) are expected to save more penalties than GK2 (18.6\%).

\setlength{\tabcolsep}{0.5em}
\begin{table}
\caption{The expected save percentage for U1 (left) and UC2 (right) for goalkeepers GK1, GK2 and GK3 following the \textbf{early educated}, \textbf{late} and \textbf{mixed} strategies.}
\begin{center}
\begin{tabular}{@{}l|lll||lll@{}}
\toprule
    & \multicolumn{3}{c||}{\textbf{\begin{tabular}[c]{@{}c@{}}Use Case 1:\\ In-game on-target kicks\end{tabular}}}                          & \multicolumn{3}{c}{\textbf{\begin{tabular}[c]{@{}c@{}}Use Case 2:\\ Shootout on-target kicks\end{tabular}}}                         \\ \midrule
    & \multicolumn{1}{c}{\begin{tabular}[c]{@{}c@{}}early\\ educated\end{tabular}} & \multicolumn{1}{c}{late} & \multicolumn{1}{c||}{mixed} & \multicolumn{1}{c}{\begin{tabular}[c]{@{}c@{}}early\\ educated\end{tabular}} & \multicolumn{1}{c}{late} & \multicolumn{1}{c}{mixed} \\ \midrule
GK1 & 0.170                                                                        & 0.186                    & \textbf{0.193}             & 0.169                                                                        & 0.170                    & \textbf{0.177}            \\
GK2 & \textbf{0.186}                                                               & -                        & -                          & \textbf{0.187}                                                               & -                        & -                         \\
GK3 & 0.153                                                                        & \textbf{0.194}           & 0.191                      & 0.150                                                                        & 0.175                    & \textbf{0.178}            \\ \bottomrule
\end{tabular}
\end{center}
\label{tbl:use-cases}
\end{table}

\paragraph{UC2: Which goalkeeper is best for penalty shootouts?}
When participating in a tournament, a team needs to select three goalkeepers. Ideally, one of them is a strong penalty stopper that could be used in a shootout. This analysis only considers shootout penalty kicks. Table \ref{tbl:use-cases} shows the results. Interestingly, for shootout penalty kicks, GK2 now has the highest expected save percentage amongst the three goalkeepers. Following the \textbf{early educated} policy, GK2 is expected to stop 18.4\% of the kicks aimed on target. Both for GK1 and GK3 the optimal policy is now to follow the \textbf{mixed educated} policy which will let them save 17.4\% (GK1) and 17.5\% (GK2) of the penalties.

GK2 scores best in shootout kicks which might be related to the fact that in shootouts players are less likely to take their kicks keeper-dependently (21.2\% in-game vs 15.1\% in shootouts). In shootouts, there are fewer keeper-dependent kicks. Hence, the fact that GK2 cannot dive late is less important than their higher early dive range. 

\paragraph{UC3: How can the framework be used in a game?}
One potential real-life scenario is to provide a list of taker-specific penalty advice to a goalkeeper in a specific game situation, similar to the list Jordan Pickford had at Euro 2024 that was mentioned in the introduction.
Providing such advice requires knowing a goalkeeper's traits. First, it requires the early and late dive ranges, and the correct corner guess percentage when diving late. These values are typically collected by teams themselves from experiments in training and from game data if the goalkeeper has faced a sufficiently large number of penalties. Next, when facing a particular penalty taker at a specific moment of the match\footnote{The models include game state features; these could also be omitted to pre-compute probabilities.}, the framework will provide the probabilities of saving the kick for each of the policies. Sampling (rolling a weighted dice) from these probabilities will then determine what policy to use. This yields instructions that are tailored to the goalkeeper's traits. Research \cite{ibrahim2018} has shown that there is no significant difference in diving toward the right versus left side for elite goalkeepers. Furthermore, following such an instruction will remove the pressure from the goalkeeper as it will not be the goalkeeper's fault if the wrong choice was made. This will avoid the well-known action bias when goalkeepers face penalty kicks \cite{bareli2007}.

\section{Related work}

Multiple game theoretic analyses have modeled an individual penalty kick as a 2-player zero sum game where each player has two actions: shooting or diving to the natural or non-natural side~\cite{azarbareli2011,chiappori2002,palacioshuerta2003,tuyls2021}.  Palacios-Huerta \cite{palacioshuerta2003} analyzed 1,417 penalty kicks taken between 1995-2000 and showed that kickers and goalkeepers follow a mixed-strategy Nash equilibrium. Tuyls et al. \cite{tuyls2021} confirmed this on dataset containing 12,399 penalty kicks taken between 2011 and 2017. Moreover, they also performed a more fine-grained analysis where players are clustered by playing style. Their findings indicate that Nash-derived recommendations depend on players' style of play.

Sports science and performance analysis studies have investigated placement from the perspective of a kicker. Here, multiple studies analyzed between 300-1000 penalty kicks from top-level games and concluded that kicks should be directed high, ideally toward the top corners \cite{horn2020,almeida2016,bareliazar2009}. Still, there are many low kicks in penalty shootouts. 
Hunter et al. \cite{hunter2022} analyzed a small number of kicks taken by amateur and semipro players in a controlled setting. They used a hand-crafted model to analyze a shooter's choice of aim, kicking technique, and speed when taking a keeper-independent kick. It is unclear how well this would translate to game environments and professional players. Finally, several approaches have tried predicting the direction of a penalty kick~\cite{busca2022,bransen2021predicting,secco2023prediction}.

\section{Conclusion}
We analyzed a large dataset of nearly 8,000 penalties that were annotated by penalty experts. To be able to take into account action capacities, we proposed a novel simulation framework to evaluate the efficacy of different policies that goalkeepers could employ. These policies exploited novel, player-agnostic learned models to make more informed decisions. Our results indicate that following the advice provided by the models would result in saving more penalties than using more generic information. We showed how the best policy depends on the goalkeeper's capacities, that aligning slightly off center increases the expected save percentage, and that diving late is beneficial if you expect the kick to be within reach. We also discussed several practical use cases for our framework.

\bibliographystyle{splncs04}
\bibliography{references}

\newpage
\appendix


\section{Data}
We have access to two different datasets about penalty kicks taken during the course of the game and in shootouts: the World Cup Penalty Dataset and the Event Stream Dataset. Because they contain unique information needed for analyzing penalties, we merge the datasets.  However, this requires matching penalties between the two datasets which is not straightforward as they were collected in different ways by different organizations.  Hence, relevant information such as team names (e.g., Athletic Bilbao versus Athletic Club), player names (e.g. Sami Al Naji versus Halil Al Naji Sami), times (e.g., timezone used) and dates (e.g., swapping the order of month and day) are recorded differently in each dataset. Moreover, the datasets cover a disjoint set of matches. Next, we describe each dataset and the process used to align them.  

\subsection{World Cup Penalty Dataset}
Starting from January 1st 2012, all penalties kicked or faced by the players likely to be included in a 2022 Men's World Cup squad were annotated. This includes penalties from friendly games, cups (e.g., Champions League, FA Cup), domestic league games and international games. A trained penalty analyst annotated the penalties using a video tool and recorded for every annotated kick the person involved (either the goalkeeper or kicker), kick direction (left, center or right, and low vs high), dive direction (left, center or right), result (goal, save or wide), score and time, shootout identifier, foot used (left or right), kicker strategy (keeper-independent or keeper-dependent) and goalkeeper timing (early or late dive). The goalkeeper's dive timing was initially not recorded and is only available for about half of the kicks. 

Soccer experts also provided a definition to annotate the pressure level of the kick. High pressure situations arise in (a) a shootout or (b) penalties taken after the 80th minute when your team is tied or trailing by a goal. Normal pressure situations arise when penalties are taken prior to the 80th minute when your team is tied or trailing by a goal. Anything else is low pressure. 

\subsection{Event Stream Dataset}
This is the standard event data that records information about all on-the-ball events. Such data is collected by commercial providers about most (semi)professional matches.
Besides containing more penalties, this dataset records different information than the World Cup dataset. It does not annotate the kicker's strategy (independent vs. dependent) or information about the goalkeeper's tactics (dive direction, early vs. late, distraction). However, it does record three important characteristics not present in the World Cup Data. Firstly, it records the end coordinates (x,y,z) of the penalty kick, either where it was stopped by the goalkeeper or where it passes the goal line (in goal or outside the posts). Secondly, it contains information about what every player did during the game (e.g., time on field, position played, actions executed). Finally, as all penalties in a game are annotated, it provides the order of penalty takers in a shootout, which is not necessarily contained in the World Cup dataset because a kick is only annotated if it involves a player likely to participate in the World Cup.

\subsection{Merging the Data}\label{app:merging_data}

We align the penalties contained in each dataset using the following three step matching procedure. 

First, we map games. We start by mapping team names.  For each pair of team names, we assign a similarity score as the maximum of the Ratfcliff-Obershelp similarity\footnote{We use the implementation in the difflib Python library.} and the Levenshtein similarity score computed as $1-\frac{\delta}{w}$ where $\delta$ is the Levenshtein distance between the two team names and $w$ is the maximum number of words in both team names. Given one team name, we find its two most similar matches and automatically map them if the similarity score is higher than 80\% for most similar one and lower than 80\% for the second most similar one. This results in matching 89\% of the teams. The remaining teams are mapped manually. Using these team mappings, we then map games based on the involved teams and the date. If there is no exact date match we allow 1 day difference or a swapped day and month. This results in 8,613 game mappings.

Secondly, we match the penalties within a game. We use the same approach as for team names, except we exploit the fact that we have mapped games to restrict the search for similar player names to those players participating in the same game. This matched all but nine players, who were mapped manually (e.g., a manually matched player was  Jorginho, who was identified as Jorge Luiz Frello Filho in one dataset).

Finally, we map penalties within the same game. We look for penalties around the same time in the game but allow for slight differences (e.g., different time zones, extra time handled differently). One challenge is when there are multiple penalty kicks from the same player in a game, which are not differentiated in the World Cup data. In these cases, we match based on the result of the penalty, the direction of the kick and foot used. This enables mapping all but nine penalties. We checked video footage to map these.

The merged data contains 7,872 penalties.


\subsection{Order of Takers}\label{app:order-experience}

\begin{figure}[t]
\includegraphics[width=6cm]{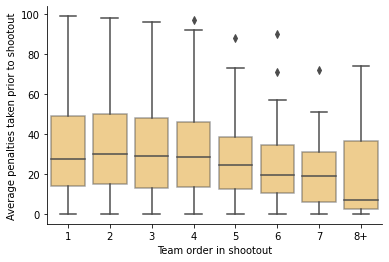}
\centering
\caption{The distribution of penalties (both shootout and non-shootout) taken by players prior to a shootout, per order in the shootout. We observe that the further we progress into the shootout, the less experienced players get.}
\label{fig:order-experience}
\end{figure}

Figure~\ref{fig:order-experience} shows the distribution on the number of prior penalties taken a player based on their position in the penalty shootout order.  While there is variance, the average number of prior kicks drops around the fifth position. Moreover, there are few takers in these positions who have previously taken a large number of kicks.  

\section{Features for the Penalty Distance and Direction Models}
\label{sec:app-features}
These features can be categorized as follows:

\begin{description}
\item[Contextual features] There are seven contextual features which record the time in match when the penalty was taken, a Boolean variable indicating shootout kicks, the goal difference (positive if kicker's team is leading), number of penalties taken in shootout, number of penalties taken by own team, and two Boolean variables indicating whether a penalty miss would lead to a loss or whether a scored penalty would lead to winning the shootout.
\item[Penalty taker and goalkeeper features] There are three general penalty taker features which record the player's preferred foot (left or right), position line (goalkeeper, defender, midfielder or striker) and age. Furthermore, we add a feature describing the height of the goalkeeper in centimeters.
\item[Penalty taker penalty experience features] There are four features describing the penalty taker's experience of taking penalties (under pressure). These are the total number of penalties taken, number of penalties scored, number of penalties taken under normal pressure and the number of penalties taken under high pressure.
\item[Penalty taker preference features] There are 18 features describing the penalty taker's potential preference for certain corners. These are the percentage of penalties to the (non-)natural corner and the percentage of penalties to the (non-)natural corner scored. The anchoring effect bias is a cognitive bias in which people tend to put a lot of weight on certain information, often being the first anchor point. Therefore we include information on the player's first penalty taken recording whether the first penalty was a goal, was saved or missed the goal, and booleans describing whether the first penalty was aimed toward the center, natural corner or non-natural corner. Similarly, people's availability bias relates to the fact that decisions are influenced by information that is directly available to a person. In our case this could be the fact that player's might be biased by their most recent penalty's result. Therefore we add the same information (goal, save or miss, and corner) for the player's most recent penalty kick.
\item[Penalty taker distance features] There are 2 features describing the player's penalty kick placement in previous penalty kicks: 1) the average distance from the center of the goal and 2) the number of kicks placed within 50 centimeters of one of the posts.
\item[Shootout features] Other information that is readily available to a player are the penalties taken by their teammates and the opponents in the ongoing shootout. These might influence their decision in their kick due to the availability bias. For this, we include 12 Boolean features. These describe the result (goal, save or miss) and the direction (center, non-natural, natural (from kicker's perspective)) of the last penalty kick taken by the opponent and by the kicker's own team. When the kicker takes the first penalty in the shootout or for the team these features values are set to NaN.
\end{description}

Some features require a player to have taken a penalty before. If this is not the case the value for features related to previous penalties are all set to NaN.


\section{Learning the uncertainty values for Equation 2}
\label{app:uncertainty}
\begin{figure}[t]
\includegraphics[width=5cm]{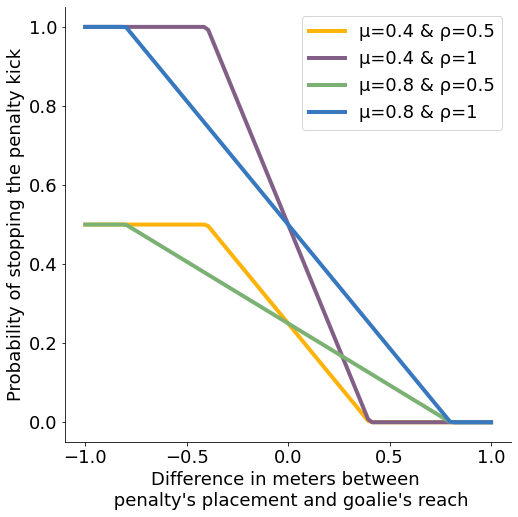}
\centering
\caption{The influence of within-reach save probability $\rho$ and uncertainty factor $\mu$ on the save probability. }
\label{fig:influence-rho-mu}
\end{figure}

Our simulation framework computes the probability of stopping penalty $k$ given policy $\pi$. This probability depends on the probability of diving toward the correct corner ($p(c|k,\pi,{gk})$) and stopping the penalty given that you are in the correct corner ($p(s|k,c,\pi,{gk})$). Equation 2 describes how we compute the probability that the goalkeeper chose the correct corner. This Equation depends on four values; 1) the Euclidean distance between the goalkeeper and the location where the ball crosses the line, 2) the goalkeeper's range following policy $\pi$, 3) the tolerance factor $\mu$, and 4) the within-reach save probability $\rho$. In this section we show that this Equation makes sense and learn the uncertainty values from our dataset.

We evaluate this Equation by comparing our predicted probability using the Equation with the actual outcome for all penalty kicks on target for which the goalkeeper was in the correct corner. We can thus view this as a probabily estimation task and hence want  our probabilities to be well calibrated. That is, if we expect the stop probability to be 0.8 for a specific group of penalties, we expect $80\%$ of these penalty kicks to be stopped by the goalkeeper.

We assume that goalkeepers stand in the center of the goal and derive the location where the ball crosses the line from our dataset. Next, we learn the most likely values for the goalkeeper's ranges for early and late dives, $\mu$ and $\rho$ using a grid search. For $\mu$ and $\pi$ we considered values in the range between 0.5 and 1 with a step size of 0.1. For $r(\pi_e)$ and $r(\pi_l)$ we considered values in the range between 2.5 and 3.5 with a step size of 0.1. The objective is to minimize the Brier score of the computed values for $p(s|k,c,\pi,{gk})$ and the actual results of the penalty kicks (saved or goal). Empirically, the values that minimize the Brier score are $r(\pi_e,{gk})=3.1$, $r(\pi_l,{gk})=2.8$, $\mu=0.7$, $\rho=0.7$. We will therefore use these values for $\mu$ and $\rho$ in our experiments. 

\begin{figure}[t]
\includegraphics[width=8cm]{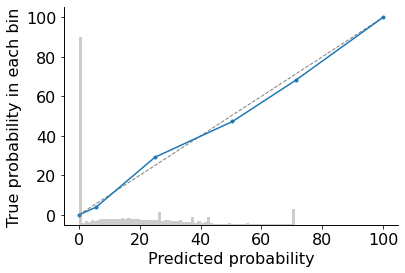}
\centering
\caption{Calibration plot for predicted probabilities of saving a penalty given that you are in the correct corner using $r(\pi_e,{gk})=3.1$, $r(\pi_l,{gk})=2.8$, $\mu=0.7$, $\rho=0.7$.}
\label{fig:calibration-simulation}
\end{figure}

Figure \ref{fig:calibration-simulation} shows the calibration plot for these settings. We observe that we quite accurately predict the save probability with Equation 2. 

\section{Model learning}
This section provides extra information on the model learning process for the Penalty Direction Model and the Penalty Distance Model. We also provide insights into what the models learned.

\subsection{Hyperparameter tuning}
\label{app:hp-set}
For both the Penalty Direction Model and the Penalty Distance Model, we tune the hyperparameters of the models using a grid search. We vary the learning rate in $\{0.01, 0.05,0.1\}$, maximum tree depth in $\{3, 4, 5, 6\}$, and the number of trees in $\{50, 100, 250\}$. Tables \ref{hyper-direction} and \ref{hyper-distance} show the optimal hyperparameters per fold.

\begin{table}
\caption{Optimal hyperparameters per fold for the Penalty Direction Model}
\begin{center}
\begin{tabular}{@{}llll@{}}
\toprule
Fold & Learning rate & Maximum depth & Number of trees \\ \midrule
1    & 0.1           & 5             & 50              \\
2    & 0.1           & 3             & 50              \\
3    & 0.1           & 4             & 50              \\
4    & 0.05          & 4             & 100             \\
5    & 0.05          & 4             & 100             \\ \bottomrule
\end{tabular}
\label{hyper-direction}
    
\end{center}
\end{table}

\begin{table}
\caption{Optimal hyperparameters per fold for the Penalty Distance Model}
\begin{center}
\begin{tabular}{@{}llll@{}}
\toprule
Fold & Learning rate & Maximum depth & Number of trees \\ \midrule
1    & 0.1           & 4             & 100             \\
2    & 0.1           & 4             & 100             \\
3    & 0.1           & 4             & 100             \\
4    & 0.1           & 4             & 100             \\
5    & 0.1           & 4             & 100             \\ \bottomrule
\end{tabular}
\end{center}
\label{hyper-distance}
\end{table}

\subsection{Calibration plots for the Penalty Direction Model}
\label{app:results}

\begin{figure}[t]
\includegraphics[width=8cm]{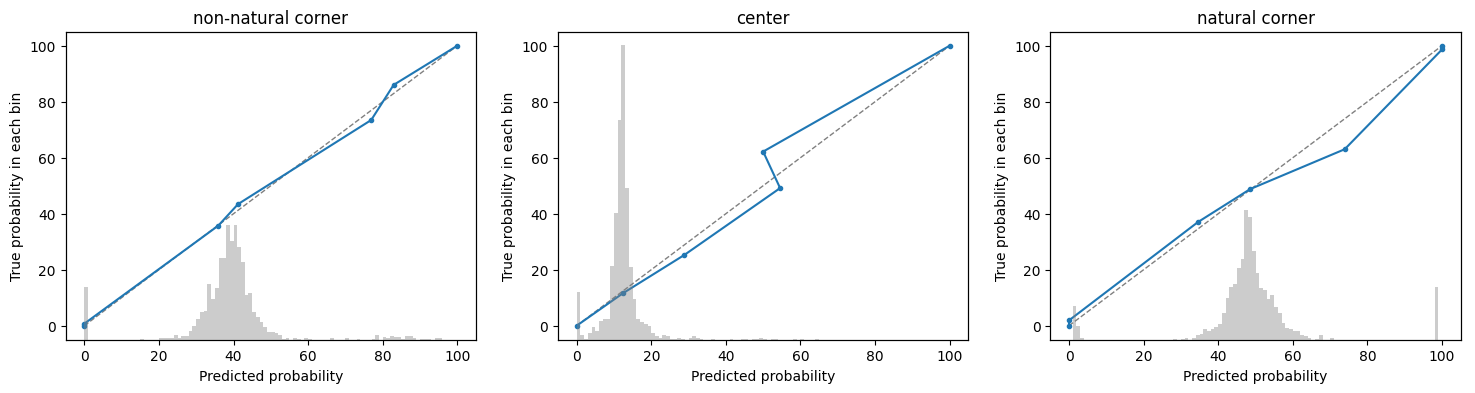}
\centering
\caption{Calibration plots for the predictions by the Penalty Direction Model for each of our classes (non-natural, center, natural).}
\label{fig:direction-calibration}
\end{figure}

Figure \ref{fig:direction-calibration} shows calibration plots for each of our classes for the Penalty Direction Model. These show that our classifier is well calibrated for all three classes.

\subsection{Insights in feature combinations for learned models}
\label{app:feature-insights}
We inspect the predictions that our Penalty Direction Model made on the dataset and inspect predictions made for different feature combinations. We especially focus on the predictions for the natural corner as that is the most common corner. We observe that players  aim for their natural corner less often when (a) they take their first ever penalty, (b) when a miss will result in losing a shootout, or (c) the previous kick from the player's team was saved in the player's natural corner, with this effect being stronger when a goal will result in a shootout win. Finally, the first kick for each team in a shootout is aimed less often to the natural corner, whereas a team's 5th kick is aimed more toward the natural corner, especially if the taker is a forward.

We also inspect the estimations that our Penalty Distance Model made for different feature combinations. We observe that the importance of the kick in the shootout and the player's penalty experience impact the player's kick placement. The model learned that players who happen to take their first penalty kick in an official match ever in a penalty shootout are more likely to place the ball closer to the goalkeeper when a goal will result in a win. Such players might be afraid to shoot wide and therefore place the ball closer to the goalkeeper in such a situation. On the other hand, experienced pressure penalty takers (at least 5 penalties taken under normal or high pressure), seem not to be influenced by this and will place the ball further away from the goalkeeper if a goal will result in a win.

\section{Reproducibility}

There are several public datasets for event data containing many penalty kicks, for example those provided by StatsBomb\footnote{https://github.com/statsbomb/open-data} and Wyscout\footnote{https://footballdata.wyscout.com/}.

StatsBomb's open source data contains 71 seasons of data across 18 different competitions. These include matches in domestic leagues like the Spanish La Liga (highest level in Spain) and the English Women's Super League (highest level in England), as well as cup games in the men's UEFA Champions League, women's UEFA Champions League, men's FIFA World Cup, Women's World Cup, FIFA U20 World Cup. These also include data for penalty shootouts in those tournaments. This dataset contains thousands of penalty kicks.

The Wyscout dataset contains data for games played in the 2017/2018 season in the top 5 domestic leagues in men's football, as well as data from the 2018 men's FIFA World Cup and the 2016 men's UEFA Euro's. This dataset contains hundreds of penalty kicks.

The annotated dataset cannot be shared for commercial reasons. However, this dataset can be reproduced by manually annotating penalties for which you annotate the following for the penalty kicks in your event dataset that you want to analyze:

\begin{enumerate}
    \item kicker strategy (independent versus dependent kick)
    \item goalkeeper dive direction (left, center, or right)
    \item goalkeeper timing (early versus late dive)
\end{enumerate}
Annotating this requires trained analysts who can identify the differences between the strategies. However, Sections 2 and 3 contain useful aggregate information about kicker strategy and dive timing.


\end{document}